\documentclass[12pt,a4paper]{cibb}

\usepackage{subfigure,graphicx}
\usepackage{amsmath,amsfonts,latexsym,amssymb,euscript,xr}
\usepackage{booktabs}
\usepackage[nodayofweek]{datetime}
\usepackage{hyperref}

\usepackage[table]{xcolor}
\usepackage{color,colortbl,tabularx}
\usepackage[most]{tcolorbox}

\usepackage[english]{babel}
\usepackage[protrusion=true,expansion=true]{microtype}
\usepackage{amsmath,amsfonts,amsthm}
\usepackage{pifont}

\definecolor{LightBlue}{rgb}{0.88,0.9,0.9}

\title{\Large $\ $\\ \bf Synthetic medical data generation: state of the art and application to trauma mechanism classification}

\author{\large Océane Doremus$^{1,\dagger}$, Ariel Guerra-Adames$^{1,2,\dagger}$, Marta Avalos-Fernandez$^{2}$, Vianney Jouhet$^{3}$, Cédric Gil-Jardiné$^{3}$, Emmanuel Lagarde$^{1,4,*}$}
\address{\footnotesize $\ $\\$^1$ AHeaD Team, Université de Bordeaux, INSERM, BPH, U1219, F-33000 Bordeaux, France. oceane.doremus@u-bordeaux.fr, 0009-0000-8361-1103\\
$^2$ SISTM Team, Université de Bordeaux, INSERM, INRIA, BPH, U1219, F-33000 Bordeaux, France. ariel.guerra-adames@u-bordeaux.fr, 0000-0002-7881-8246 \\
$^3$ CHU de Bordeaux, INSERM, U1219, F-33000 Bordeaux, France. vianney.jouhet@u-bordeaux.fr, 0000-0001-5272-2265 \\
$^4$ Aix-Marseille Université, F-13015 Marseille, France. emmanuel.lagarde@u-bordeaux.fr, 0000-0001-8031-7400 \\
\bigskip
$^\dagger$ These authors contributed equally to this study. \\
$^*$ Corresponding author
}

\abstract{\small synthetic data generation, electronic health records, variational autoencoders, diffusion models, large language models. \normalsize
\\[17pt]
{\bf Abstract.} Faced with the challenges of patient confidentiality and scientific reproducibility, research on machine learning for health is turning towards the conception of synthetic medical databases. This article presents a brief overview of state-of-the-art machine learning methods for generating synthetic tabular and textual data, focusing their application to the automatic classification of trauma mechanisms, followed by our proposed methodology for generating high-quality, synthetic medical records combining tabular and unstructured text data. }

\begin{document}
\thispagestyle{myheadings}
\pagestyle{myheadings}
\markright{\tt Proceedings of CIBB 2025}%check year

\section{\bf Introduction}
\label{sec:INTRODUCTION}

Research in machine learning for health faces a major challenge of reproducibility, a concept that refers to the ability to systematically obtain the same results as those initially reported when the original data and code are available. McDermott et al. \cite{mcdermott2021reproducibility} introduce in this regard the notion of \textit{frictionless reproducibility}, defined as the ability to easily reproduce results through transparent data sharing, the availability of easily re-executable code, and rigorous evaluation using standardized benchmarks. In their review of over 500 studies, the authors find that many published works in machine learning for health do not meet these criteria, particularly due to the constraints involved in sharing patient data because of its sensitive nature.

This lack of reproducibility is often justified by several factors, such as the inherent sensitive nature of health data or even the implementation of governance policies like the AI Act in Europe, which classifies certain AI systems for health as ``high-risk''\footnotemark due to their capacity to influence critical decisions concerning human health. Additionally, the General Data Protection Regulation (GDPR) and the recommendations of the CNIL (French Data Protection Authority) impose further constraints on the use and sharing of health data\footnotemark, thereby reinforcing these reproducibility challenges

\footnotetext[3]{Regulation (EU) 2024/1689 laying down harmonised rules on artificial intelligence (AI Act).}
\footnotetext[4]{Regulation (EU) 2016/679 of the European Parliament and of the Council of 27 April 2016 on the protection of natural persons with regard to the processing of personal data and on the free movement of such data (General Data Protection Regulation – GDPR).}

This problem is exacerbated by the heterogeneity of data collection systems in healthcare institutions, combined with biases introduced by local practices. For example, differences in how medical records are structured or coded between institutions can lead to difficulties in reproducing results obtained in a specific region or context \cite{wiens2014study}. This challenge is particularly notable in classification tasks applied to medical conditions, where the sensitivity of patient data and the rarity of certain specific cases can limit the creation of robust and diverse training datasets.

These constraints raise a fundamental question: Is so-called \textit{frictionless reproducibility} compatible with machine learning in health? The use of synthetic data could provide an answer to this question. This artificially generated data, mimicking the statistical properties of real data, could offer an alternative to overcome obstacles related to confidentiality and accessibility, while allowing the development and evaluation of reproducible learning algorithms.

However, this approach, in turn, raises two main questions that we address in our study: \textit{1) Can we generate \textbf{high-quality} synthetic medical data that is respectful of \textbf{patient privacy}} and \textit{2) How can we \textbf{evaluate} these two aspects?}

To answer these questions, we plan to implement and evaluate methods for generating synthetic tabular and textual medical data, with the aim of training high-performing models for the classification of trauma mechanisms. As a first step, we have conducted a literature review on methodologies for generating synthetic data adapted to tabular medical data and unstructured texts (Section \ref{sec:SOTA}), as well as on approaches for evaluating their quality and respect for confidentiality (Section \ref{sec:Methodology}).

\section{\bf State of the art}
\label{sec:SOTA}

The practical value of synthetic data depends heavily on the sophistication of its generation methods and thorough evaluation of its fidelity, utility for specific tasks, and the robustness of its privacy safeguards. To delineate the current advancements and methodologies, we conducted a scoping review utilizing the Scopus, PubMed, and Google Scholar academic databases, targeting publications on the creation or assessment of synthetic medical data. This review identified 164 articles, based on which we will outline the predominant generation techniques and the frameworks for their evaluation.

\paragraph{Generation Methods. }
Historically, in statistics, the generation of simulated data has been a key step in evaluating and validating theoretical models, allowing hypotheses to be tested under controlled conditions. This approach relies on probability distributions and parametric models to simulate appropriate scenarios. In contrast, in the field of machine learning, the focus has instead been on evaluating algorithms using real datasets, often well-documented and standardized, to test their ability to generalize to concrete situations \cite{breiman2001statistical}.
Recently, synthetic data generation has undergone significant evolution, moving from purely statistical techniques to deep learning approaches \cite{budu2024evaluation}. Statistical approaches include random sampling and perturbation of existing data, as well as decision trees \cite{vaidya2018scalable} and Bayesian networks \cite{kaur2021application}.
Variational autoencoders (VAEs) \cite{kingma2013auto} extended the probabilistic approach by combining Bayesian inference with deep neural networks. Generative adversarial networks (GANs) \cite{goodfellow2014}, on the other hand, introduced a principle of opposition between two networks to improve the quality of generated data. However, GANs are known for their instability \cite{abedi2022gan}.
Diffusion models, which are more recent, operate by progressively adding noise to real data and then learning to reverse this process \cite{kotelnikov2023tabddpm}. For textual data, large language models (LLMs) offer interesting prospects, although more complex approaches are needed to adapt them correctly to the medical domain \cite{Kumichev2024}.

\paragraph{Evaluation of Generated Data. }
The evaluation of synthetic data remains complex in the absence of established standards; however, the literature identifies several broad categories of metrics for assessing various aspects. The quality of synthetic data primarily revolves around two axes: fidelity and utility \cite{budu2024evaluation}. Fidelity measures how closely the generated data resembles the original data, using measures such as the Wasserstein distance or the Kolmogorov-Smirnov test. Utility, on the other hand, assesses the relevance of synthetic data for specific applications, notably by measuring performance on concrete tasks, such as classification.
Beyond these aspects, privacy metrics aim to ensure that synthetic data do not allow for the reconstruction or identification of individuals present in the original dataset \cite{yan2022multifaceted}. Finally, the notion of equity or \textit{fairness} is also gaining importance, seeking to ensure that generated data do not perpetuate or amplify biases present in the original data \cite{vallevik2024can}.

\paragraph{Balance between Privacy and Utility. }
The balance between the confidentiality and utility of synthetic data constitutes a major challenge, as their mutual reinforcement is often antagonistic \cite{Stadler2020PRIVACYMIRAGE}; hence the need for rigorous oversight of their generation and use to ensure accountability and security.
\section{\bf Classification of trauma mechanism from emergency data}
A detailed understanding of trauma mechanisms is a major challenge for public health and prevention efforts. In this context, our objective is to enhance the understanding and surveillance of trauma in France by developing models for the automated production of epidemiological indicators related to trauma mechanisms, using data collected in emergency departments.

We use free-text clinical notes from electronic health records written during patient check-in at an emergency department from the Bordeaux University Hospital. Theses notes written by the medical/nursing staff describe in details circumstances and mechanisms of the trauma event. These notes are a rich source of information, but they also present challenges for training robust classification models. Each hospital has its own documentation practices, leading to significant heterogeneity in the style, structure, and level of detail in the notes. Moreover, the diversity of trauma mechanisms-ranging from domestic falls to road traffic accidents, as well as assaults or work related injuries, makes data processing more complex, especially since some mechanisms are rare and therefore under-represented in the training datasets. 

According to WHO recommendations, five main epidemiological categories of trauma are generally distinguished: occupational accident, accident of daily life, road traffic accident, assault, and self-harm. This classification can be further refined according to the location, context, or specific mechanism of the trauma (e.g., fall, collision), which increases the complexity of the analysis and accentuates class imbalance. A previous study based on data from the Bordeaux University Hospital \cite{chenais}, showed a significant class imbalance problem, negatively impacting the accuracy of automatic trauma classification algorithms. Some classes in the aforementioned dataset were represented by less than 2.5\% of the total sample size (eg. intentional injury).

The main methodological challenge is therefore to train models that are both effective and generalizable, able to adapt to the diversity of clinical contexts and heterogeneous documentation practices. This requires a sufficiently large and varied dataset that covers all trauma mechanisms, including less frequent cases. The central question remains: how can we ensure the robustness of classification models in view of this diversity and class imbalance?

\section{\bf Proposed Methodology}
\label{sec:Methodology}
Taking into account the challenges identified in the literature, we now direct our study towards a concrete application: the classification of physical trauma mechanisms.
We propose a sequential hybrid approach for generating synthetic tabular and textual data, with the aim of training models for the classification of trauma mechanisms from emergency department patient records. The objective is to faithfully recreate the distributions of our real data while respecting privacy, and also to enrich the training and evaluation datasets, particularly for rare trauma mechanisms, in order to improve the robustness of future classification models. Finally, from this synthetic tabular data, we will then generate unstructured clinical texts that simulate plausible medical histories.

\begin{figure}[h]
 \begin{center}
 \includegraphics[width=0.99\textwidth]{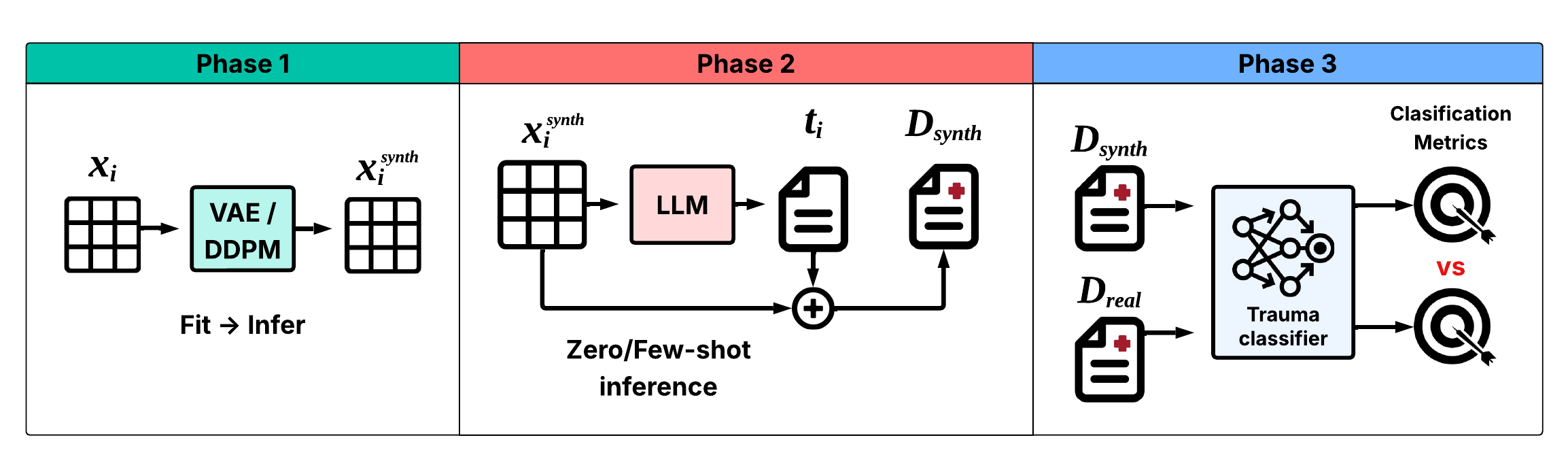} \caption{Proposed 3-phase methodology}
 \label{fig:workflow}
 \end{center}
 \vspace{-8mm}
\end{figure}

This methodology will unfold in three phases:

\paragraph{Phase 1: Generation of tabular data. }
We will train two different deep generative models incorporating differential privacy guarantees \cite{dwork2006differential}, notably a VAE and a denoising diffusion probabilistic model (DDPM). These models will be trained on a real database from the adult emergency department of the Pellegrin site of the Bordeaux University Hospital, comprising over 700,000 samples collected between 2013 and 2024. Each sample \( x \in \mathbb{R}^{d} \) consists of \(d=25\)  categorical and continuous variables, including patient vital signs (heart rate, blood pressure, etc.), contextual information about the emergency visit (arrival time, admission mode, etc.), as well as demographic data (age, sex, etc.).
Formally, we seek to learn a synthetic distribution \( p_{\theta}(x) \) approximating the unknown real distribution \( p_{\text{r}}(x) \), while respecting a strict constraint of Rényi differential privacy \cite{mironov2017renyi}. Once trained, the model will allow for the sampling of a new synthetic dataset: 
\begin{equation}
    X_{\text{synth}} = { x_i^{\text{synth}} } \quad {i=1}^{N}, \quad x_i^{\text{synth}} \sim p{\theta}(x).
\end{equation}

To improve class balance, we will intentionally generate a larger number of samples corresponding to rare or underrepresented traumatic mechanisms in the initial real data. This strategy will allow us to obtain a balanced synthetic dataset facilitating the effective training of predictive models. 

\paragraph{Phase 2: Automatic generation of free text. }
The synthetic tabular data \( X_{\text{synth}} \) will serve as input to the pre-trained model MedGemma 27B, optimized for deep understanding of medical text and clinical reasoning, used in a \textit{zero-shot} or \textit{few-shot} configuration.
Specifically, for each synthetic patient visit represented by the tabular variables \( x_i^{\text{synth}} \), the model will generate a realistic textual clinical note \( t_i \) from a simple instruction such as detailed in the following example :
The final set of synthetic data thus obtained will therefore be composed of both tabular and textual data: 
\begin{equation}
    D_{\text{synth}} = {(x_i^{\text{synth}}, t_i)}_{i=1}^{N}
\end{equation}
This phase is essential because narrative descriptions of the circumstances of the trauma (medical histories) contain crucial information, often not captured by tabular data alone, for an accurate classification of mechanisms. The use of LLMs to generate these texts from synthetic tabular data allows for the creation of realistic bimodal datasets.

\paragraph{Phase 3: Evaluation and Application to the classification of traumatic mechanisms. }
The synthetic tabular data will be evaluated using the metrics presented previously. Complementarily, we will also solicit emergency healthcare professionals who will qualitatively evaluate (``natural'' or ``unnatural'' a random sub-sample of (\(D_{\text{synth}}\)), composed jointly of numerical variables and associated clinical notes.
This final phase will allow us to concretely justify (or not) the utility of synthetic data for the envisaged classification task. By training trauma mechanism classification models on this data, and by comparing their performance according to the Train on Synthetic, Test on Real (TSTR) versus Train on Real, Test on Real (TRTR) paradigm, we will evaluate whether the proposed approach allows for obtaining models that are as performant, or even better, while overcoming the constraints related to the use of sensitive and sometimes imbalanced real data.

\section*{\bf Conflict of interests}
\label{sec:CONFLICT-OF-INTERESTS}
The authors declared no potential conflicts of interest with respect to the research, authorship, and/or publication of this article.

\section*{\bf Funding}
\label{sec:FUNDING}
The research efforts of Océane Doremus are supported by a doctoral grant of the \textit{École des hautes études en santé publique} (EHESP). The research efforts of Ariel Guerra-Adames are supported by a doctoral grant from the Digital Public Health Graduate Program of the University of Bordeaux (CD EUR DPH). Data collection and processing is possible with funding from the TARPON project (HDH).

\footnotesize
\bibliographystyle{unsrt}

\normalsize

%BOMBOCLAT

\end{document}